\documentclass[sigconf]{acmart}
\usepackage[T1]{fontenc}
\pdfoutput=1
\usepackage{bbm}
\usepackage{amsfonts}
\usepackage{verbatim}
\usepackage{graphicx,subfigure,multirow}
\usepackage{epstopdf}
\usepackage{comment}
\usepackage{array}
\usepackage{enumitem}
\usepackage{mathtools}
\usepackage{amsmath}
\usepackage{color}
\usepackage{multirow}
\usepackage{balance}
\usepackage{url}
\usepackage{eurosym}
\usepackage{booktabs} 
\usepackage{xr}
\usepackage{algpseudocode}

\makeatletter
\newif\if@restonecol
\makeatother

\usepackage{algorithm}
\usepackage{listings}

\pdfoutput=1
\usepackage{pdfpages}
\usepackage{etoolbox}
\makeatletter
\AfterEndEnvironment{algorithm}{\let\@algcomment\relax}
\AtEndEnvironment{algorithm}{\kern2pt\hrule\relax\vskip3pt\@algcomment}
\let\@algcomment\relax
\newcommand\algcomment[1]{\def\@algcomment{\footnotesize#1}}
\renewcommand\fs@ruled{\def\@fs@cfont{\bfseries}\let\@fs@capt\floatc@ruled
  \def\@fs@pre{\hrule height.8pt depth0pt \kern2pt}%
  \def\@fs@post{}%
  \def\@fs@mid{\kern2pt\hrule\kern2pt}%
  \let\@fs@iftopcapt\iftrue}
\makeatother

\newcommand{\hide}[1]{} 
\newcommand{\vpara}[1]{\vspace{0.05in}\noindent\textbf{#1 }}

\newcommand{\eg}{{\sl e.g.}}
\newcommand{\ie}{{\sl i.e.}}
\newcommand{\etc}{{\sl etc.}}

\newcommand{\wrt}{{\sl w.r.t.}}

\usepackage{xspace}

\newcommand{\modelname}{MBVR}

\newcommand{\autodata}{\textbf{Auto}\xspace}
\newcommand{\mandata}{\textbf{Manual}\xspace}

\copyrightyear{2022}
\acmYear{2022}
\setcopyright{acmlicensed}\acmConference[SIGIR '22]{Proceedings of the 45th International ACM SIGIR Conference on Research and Development in Information Retrieval}{July 11--15, 2022}{Madrid, Spain}
\acmBooktitle{Proceedings of the 45th International ACM SIGIR Conference on Research and Development in Information Retrieval (SIGIR '22), July 11--15, 2022, Madrid, Spain}
\acmPrice{15.00}
\acmDOI{10.1145/3477495.3531899}
\acmISBN{978-1-4503-8732-3/22/07}

\begin{document}
\fancyhead{}
\title{Modality-Balanced Embedding for Video Retrieval}
\author{\  Xun Wang$^1$~\qquad Bingqing Ke$^1$~~\qquad Xuanping Li$^1$~~\qquad Fangyu Liu$^2$\\
Mingyu Zhang$^1$~~\qquad Xiao Liang$^1$~~\qquad Qiushi Xiao$^1$~~\qquad Cheng Luo$^1$ ~~\qquad Yue Yu$^1$\\
${}^1$~Kuaishou \quad ${}^2$~University of Cambridge}


 
 

 
 
 

\begin{abstract}
Video search has become the main routine for users to discover videos relevant to a text query on large short-video sharing platforms. 
During training a query-video bi-encoder model using online search logs,\textit{ we identify a modality bias phenomenon that the video encoder almost entirely relies on text matching, neglecting other modalities of the videos such as vision, audio, \etc} This \textbf{modality imbalance} results from a) modality gap: the relevance between a query and a video text is much easier to learn as the query is also a piece of text, with the same modality as the video text; b) data bias: most training samples can be solved solely by text matching. Here we share our practices to improve the first retrieval stage including our solution for the modality imbalance issue. We propose \textbf{\modelname} (short for Modality Balanced Video Retrieval) with two key components: manually generated \textbf{modality-shuffled (MS)} samples and a \textbf{dynamic margin (DM)} based on visual relevance. They can encourage the video encoder to pay balanced attentions to each modality. Through extensive experiments on a real world dataset, we show empirically that our method is both effective and efficient in solving modality bias problem. We have also deployed our ~\modelname~ in a large video platform and observed statistically significant boost over a highly optimized baseline in an A/B test and manual GSB evaluations. 
\end{abstract}


\maketitle

\section{Introduction}

Video search, which aims to find the relevant videos of a query from billions of videos, is essential to video-sharing platforms(\eg, TikTok, Likee, and Kuaishou). 
To be efficient, most video search systems adopt a multi-stage pipeline that gradually shrinks the number of candidates. The first stage, known as \textit{retrieval}, recalls thousands of candidates from billions efficiently, determining the upper bound of the overall performance of a search engine. The subsequent pre-ranking stages further shrink the candidates to the size of hundreds, and the final ranking server then scores and selects videos to display for users. In this paper, we focus on improving the retrieval stage of video search with multimodal embedding learning. 

With the recent development of embedding learning ~\cite{bengio2013representation} and pre-trained language models~\cite{liu2019roberta,devlin2019bert,zhan2020repbert}, embedding-based retrieval approaches have obtained promising results in web (\ie, document) retrieval~\cite{baidu_pre,Huang2013LearningDS,guu2020realm,karpukhin2020dense,hard_negs,khattab2020colbert} and product search~\cite{li2021embedding_taobao,zhang2020towards_jd}. Most of them adopt a bi-encoder architecture and are trained on labeled data or online logs.
\emph{However, when training a query-video bi-encoder with online search logs, we have identified a bothering \textbf{modality imbalance} phenomenon}: a video's embedding overly relies on its associated text contents, neglecting its visual information. Such models would falsely recall videos that only matching the query textually, but with irrelevant vision contents. Notice that on video-sharing platforms, a video is usually composed of multiple modalities including video frames, audio, text (\eg, title and hashtag), and etc. In this paper, we select two modalities to represent the video content: the text modality from the title, banners, hashtags, and the vision modality from a key frame or the cover.\footnote{The other modalities like audio, are dropped here due to the information being either extremely noisy or negligible in our scenario.} 

This modality imbalance phenomenon results from: 1) \textbf{modality gap}: both query and text are of the same textual modality, thus their relevance is easier to grasp than other video's modalities; 2) \textbf{data bias}: current search engines are mostly based on text matching, at lexical or semantic level, thus the online search logs, which are used as the training data, are heavily biased towards examples with high query-text similarities.

Recent research in video retrieval mostly focuses on designing more sophisticated architectures~\cite{videobert,lei2021less,huang2020pixel,liu2021hit_moco_video-ret} or stronger cross-modal fusion operations~\cite{mmt,qu2021dynamic,xu2021videoclip}. 
They require large-scale clean training data and heavy computational resources, making them suitable for only specific settings. What's more, in a real-world scenario with modality biased data like the video search logs, they unavoidably suffer from the modality imbalance problem. 

\begin{figure}[t]
\centering
\includegraphics[width=1.0\linewidth]{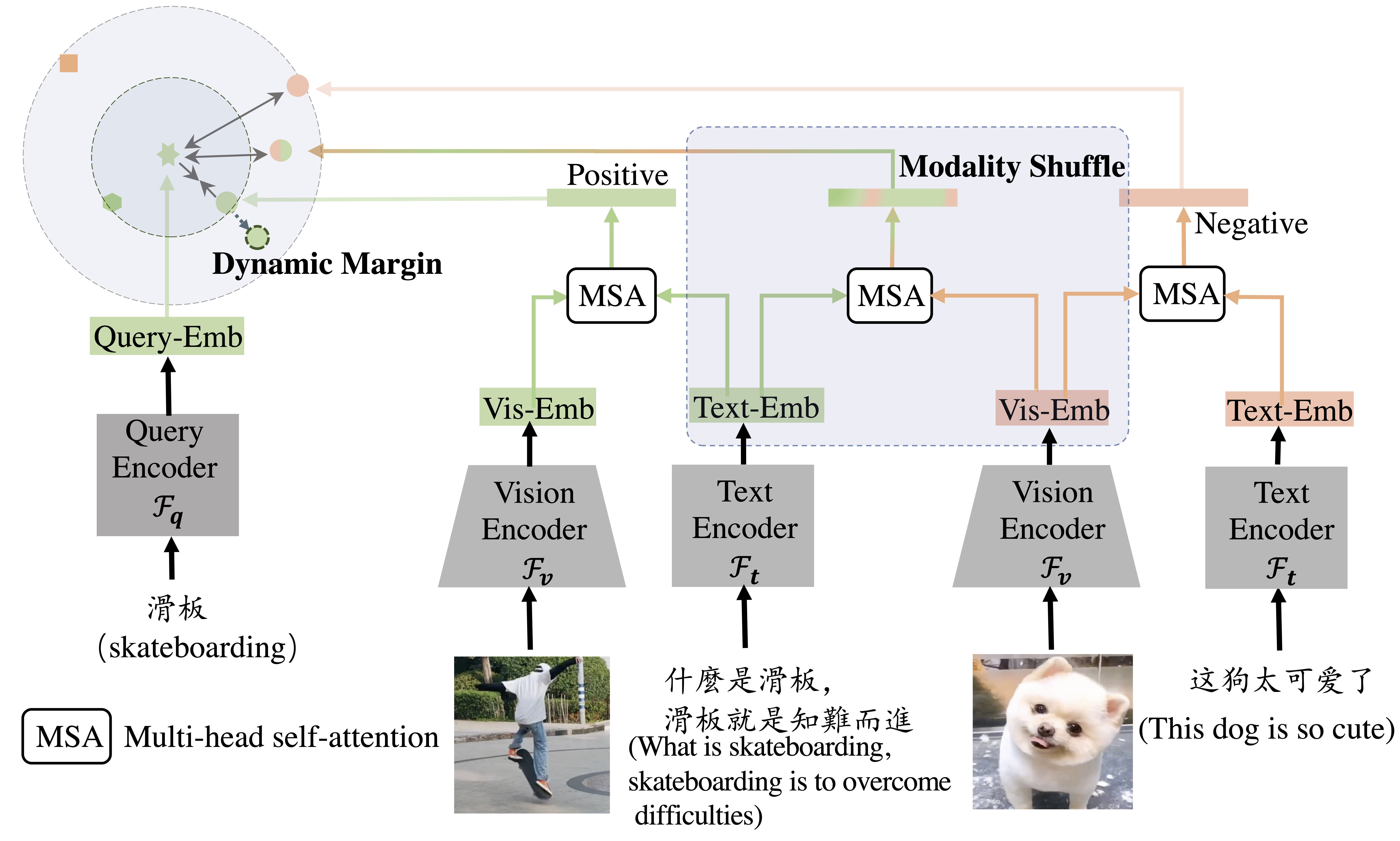}
\caption{
A graphical illustration of \modelname.
}
\vspace{-0.1in}
\label{fig:main}
\end{figure}

To bridge this gap, our paper offers a feasible solution named {\modelname} to learning modality-balanced video embeddings using noisy search logs, which is a bi-encoder framework with two key components, illustrated in Fig.~\ref{fig:main}. 

\begin{figure*}[!t]
\vspace{-0.2in}
  \centering
  \subfigure[$R_{vt}$ distribution]{
    \label{subfig:rtv}
    \includegraphics[width=0.321\textwidth]{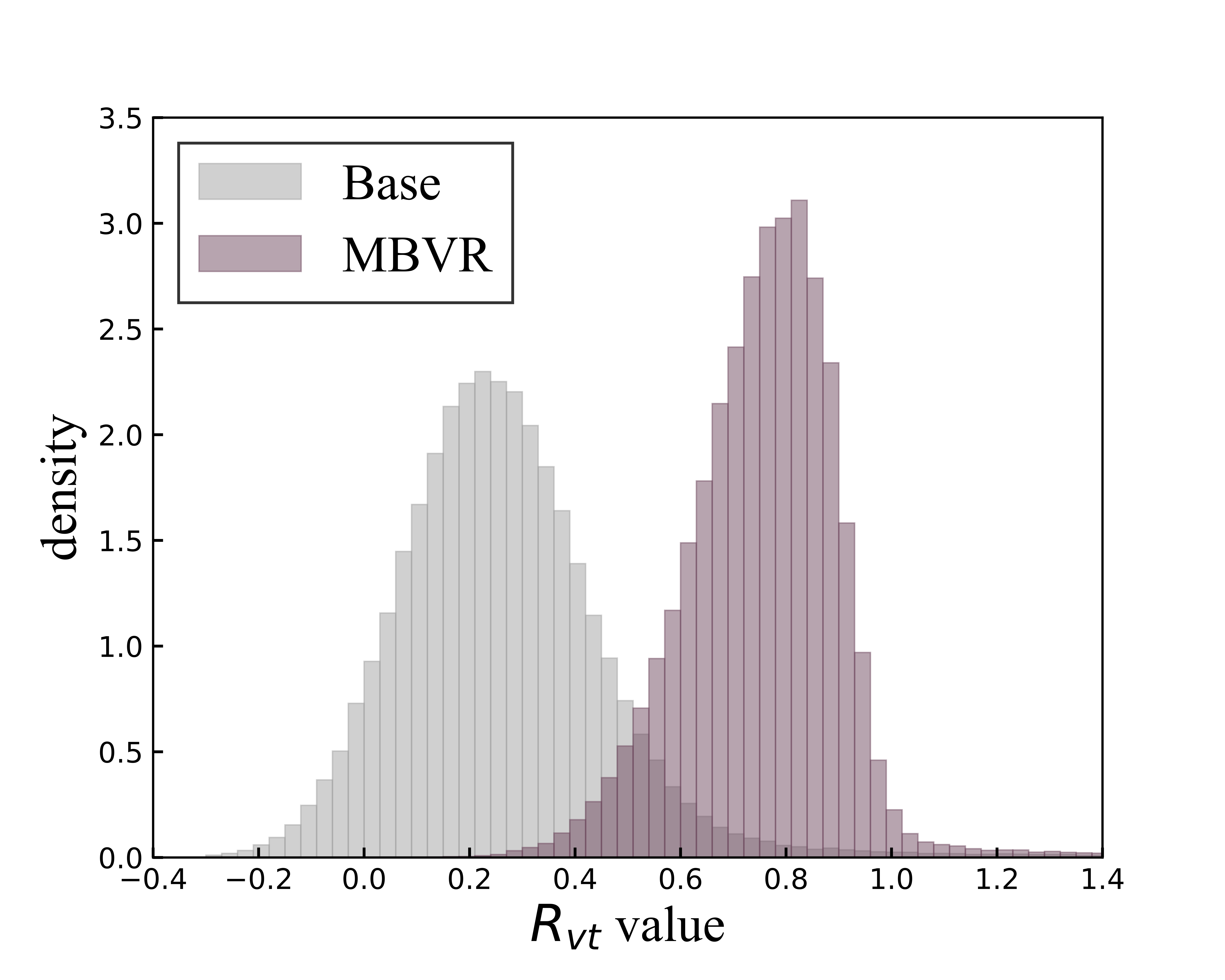}
  }
  \subfigure[Similarity scores of base model]{
    \label{subfig:sim_density_bi}
    \includegraphics[width=0.321\textwidth]{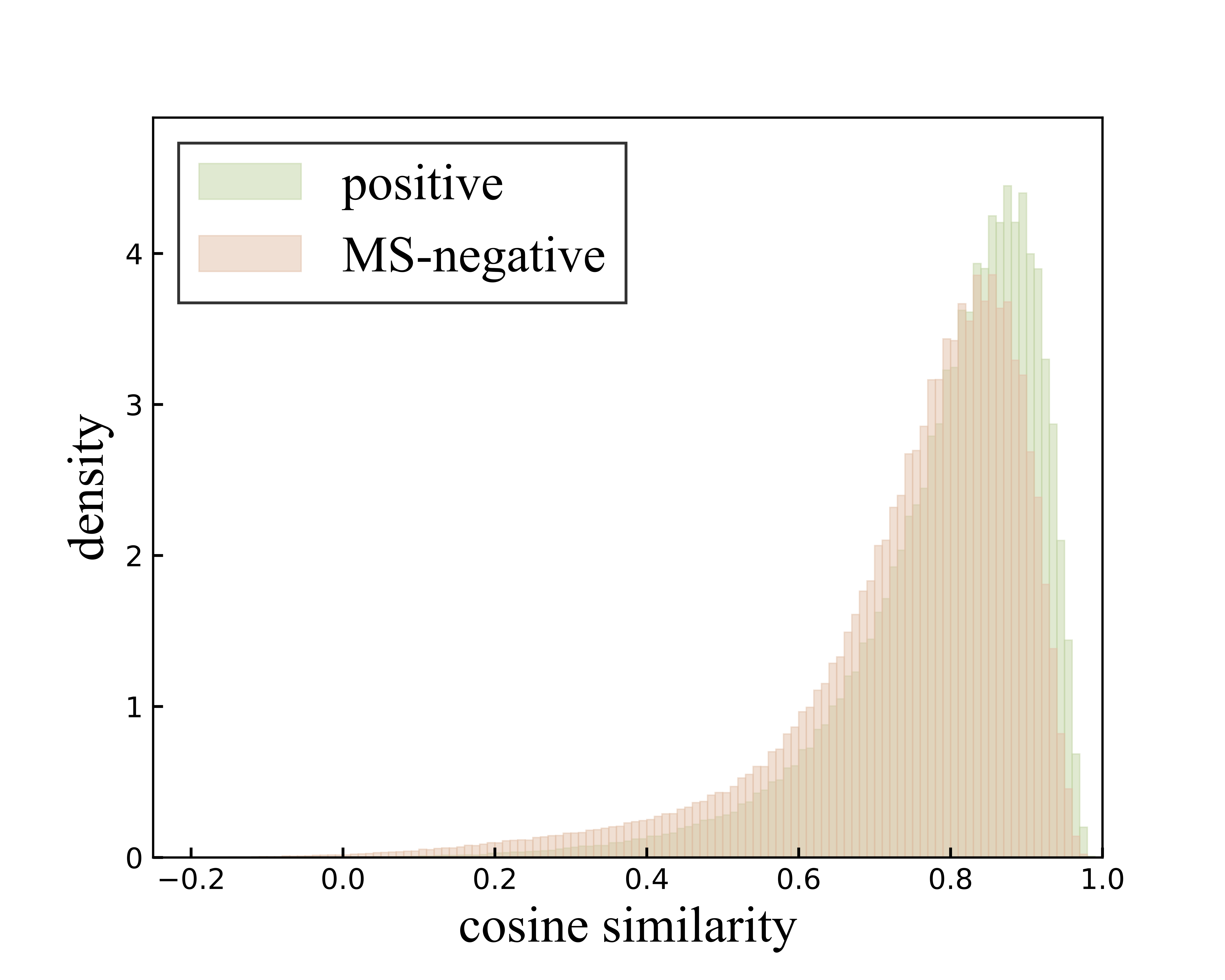}
  }
  \subfigure[Similarity scores of MBVR]{
    \label{subfig:sim_density}
    \includegraphics[width=0.321\textwidth]{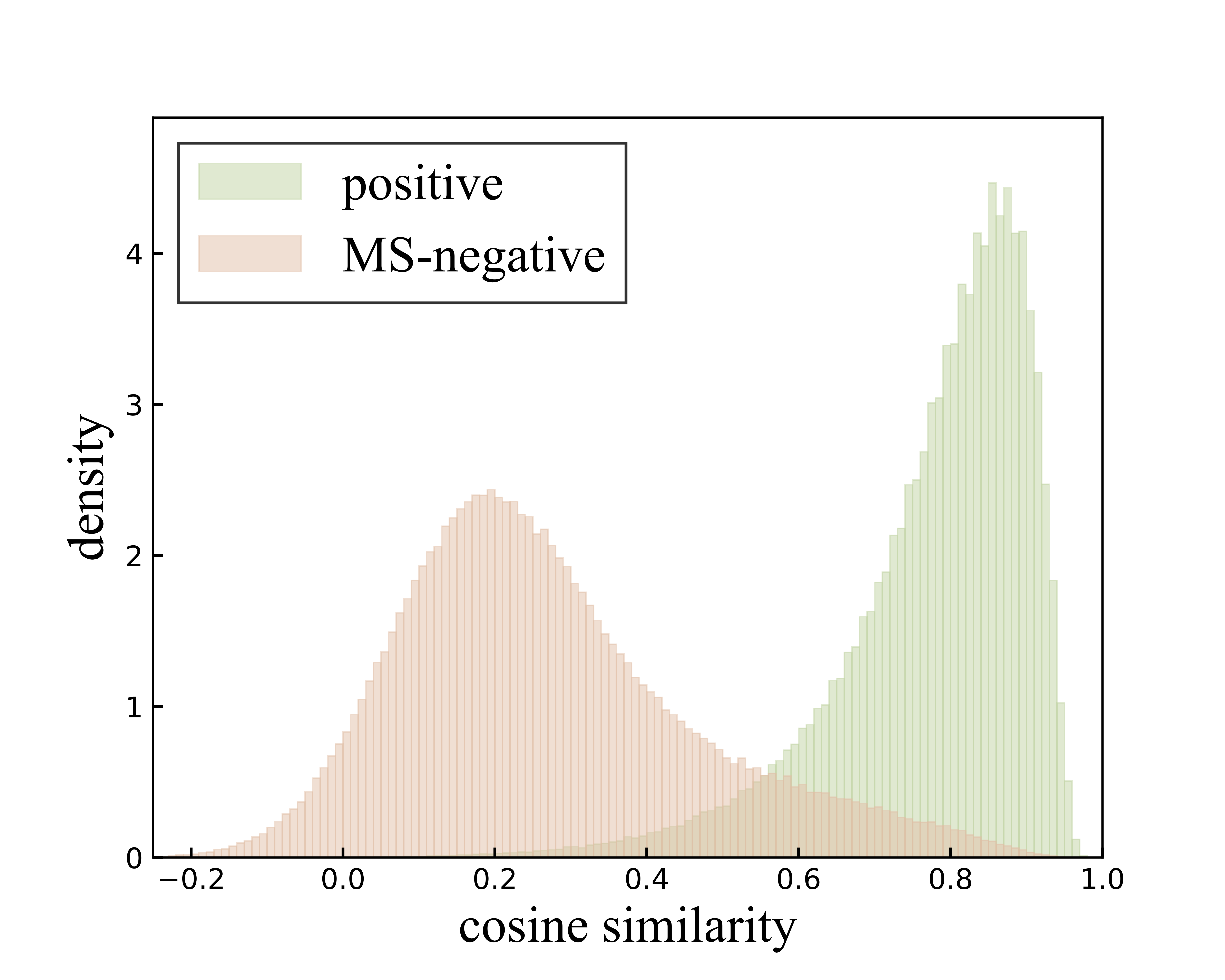}
  }
  \vspace{-0.1in}
  \caption{(a) $R_{vt}$ (the ratio of the vision modality influence to the text modality influence) distribution of base model and MBVR. (b) Similarity scores between the queries and the positives/MS negatives of base model and (c) that of MBVR. 
  \label{fig:sim_den_shuffle_neg}
  }
\end{figure*}

\vpara{Modality-Shuffled negatives.}
To correct the modality imbalance bias, we generate novel \textit{modality-shuffled (\textbf{MS}) negative samples} that train the model adversarially. An MS negative consists of a relevant text and an irrelevant video frame~\wrt~a query. MS negatives can be mistakenly ranked at top if a model overly relies on a single modality (in Fig.~\ref{subfig:sim_density_bi}). We add an additional objective to explicitly punish wrongly ranked MS negatives.  

\vpara{Dynamic margin.} We further enhance the model with a margin dynamically changed \wrt~ the visual relevance. The dynamic margin amplifies the loss for the positive query-video pairs that with both related texts and vision contents. Thus, the models with dynamic margin pull visually relevant videos closer to the query.  

We conduct extensive offline experiments and ablation study on the key components to validate the effectiveness of \modelname~over a strong baseline and recent methods related to modality balancing~\cite{kim2020modality,lamb2019interpolated}. Furthermore, we deploy an online A/B test and GSB evaluations on a large video sharing platform to show that our \modelname~improves the relevance level and users' satisfaction of the video search.

\section{\modelname}
\label{sec:model}
In this section, we first introduce the model architecture, the training of a strong baseline model. Then we illustrate the modality imbalance issue with statistical analysis, and introduce our \modelname~ with generated negatives and the dynamic margin. 

\subsection{Model Architecture}
\label{subsec:arch} 
Our model architecture follows the popular two-tower (\ie, bi-encoder) formulation, as \cite{Huang2013LearningDS,li2021embedding_taobao,zhang2020towards_jd,baidu_pre,hard_negs,liu2021que2search}, with a transformer-based text encoder of query and a multi-modal encoder of video.

\vpara{Query encoder $\mathcal{F}_q$} can be summarised as \verb|RBT3+average+FC|.
We utilize the RoBERTa~\cite{liu2019roberta} model with three transformer layers as the backbone \footnote{We have also tried to use larger bert encoders with 6 and 12 transformer layers. However, such large models only bring negligible effect with much heavier computational cost, thus we choose use the 3-layers RoBERTa as our text encoder.} and use \verb|average+FC| (\ie, average pooling followed by a fully connected layer) to compress the final token embeddings to $d-$dim ($d=64$). 

\vpara{Multimodal encoder $\mathcal{F}_m$} consists of a text encoder $\mathcal{F}_t$, a vision encoder $\mathcal{F}_v$ and \textbf{a fusion module} $\mathcal{H}$. For a video sample $m$, its embedding is computed as
\begin{align}
    \mathcal{F}_m(m) =\mathcal{F}_m(t, v) = \mathcal{H}(\mathcal{F}_t(t),~\mathcal{F}_v(v)),  \label{eq:video_encoder}
\end{align}
where $t$ is the text input and $v$ is the vision input. The text encoder $\mathcal{F}_t$ shares weights with $\mathcal{F}_q$ \footnote{Such weight sharing brings several benefits,~\eg, reducing model parameters to save memory and computational cost, and introducing prior query-text matching relation to regularize the training~\cite{firat2016zero,xia2018model,baidu_pre}.}. The vision encoder $\mathcal{F}_v$ adopts the classical ResNet-50~\cite{he2016resnet} network. For the fusion module $\mathcal{H}$, we adopt the multi-head self-attention (MSA)~\cite{vaswani2017attention} to dynamically integrate the two modalities and aggregate the outputs of MSA with an average pooling. 
We have also tried other feature fusion operations (\eg, direct addition and concatenation-MLP) and discovered that MSA works the best.  

\subsection{Base Model}
\label{subsec:main}
Most existing works,~\eg, \cite{Huang2013LearningDS,baidu_pre,huang2020_facebook_embedding}, train bi-encoders with the approximated query-to-document retrieval objective. Specifically, given a query $q$, its relevant videos $\mathcal{M}^+_q$, and its irrelevant videos  $\mathcal{M}^-_q$, the query-to-document objective is as below:
\begin{align}
    \mathcal{L}_{qm} = -\log\Big(\frac{\exp(s(q, m)/\tau)}{\exp(s(q, m)/\tau)+\sum_{\hat{m}\in M^-_q}\exp(s(q, \hat{m})/\tau)}\Big), \label{eq:qm}
\end{align}
where $\tau$ is a temperature hyper-parameter set to 0.07, and $s(q, m)$ is the cosine similarity of a query-video pair $(q, m)$ (\ie, $s(q, m)=cos<\mathcal{F}_{q}(q), \mathcal{F}_{m}(m)>$). Notably, here we adopt in-batch random negative sampling, which means $\mathcal{M}^-_q$ is all the videos in current mini-batch except the positive sample $m$.

As recent works \cite{liu2021que2search,liu2021hit_moco_video-ret} added a conversed document-to-query retrieval loss, we formulate a corresponding  $\mathcal{L}_{mq}$ as below:

\begin{align}
    \mathcal{L}_{mq}= -\log\Big(\frac{\exp(s(q, m)/\tau)}{\exp(s(q, m)/\tau)+\sum_{\hat{q}\in\mathcal{Q}^-_m}\exp(s(\hat{q}, m)/\tau)}\Big), \label{eq:mq}
\end{align}
where $\mathcal{Q}^-_m$ denotes the irrelevant queries of the video $m$, ~\ie, all the queries in current mini-batch except $q$.  

The sum of the query-to-document loss and the reversed document-to-query loss results in the main bidirectional objective: 
\begin{align}
    \mathcal{L}_{bi} = \mathcal{L}_{qm} + \mathcal{L}_{mq}. \label{eq:bi}
\end{align}

Beside the above bidirectional objective optimizing the relevance between the query embedding and the video's multimodal embedding, we also add a auxiliary task $\mathcal{L}_{t}$ ($\mathcal{L}_{v}$) to optimize the relevance between the query and the video's text modality (the vision modality) with similar formulations. The whole objective $\mathcal{L}_{base}$ for the base model is:
\begin{align}
    \mathcal{L}_{base} = \mathcal{L}_{bi} + \alpha\mathcal{L}_{v} +  \beta\mathcal{L}_{t}, \label{eq:base}
\end{align}
where $\alpha=\beta=0.1$, are the weight hyper-parameters. 

\subsection{Statistical Analysis of Modality Imbalance}
\label{subsec:stat}
To identify the modality imbalance, we define an indicator ${R}_{vt}$ as below, 
\begin{align}
    {R}_{vt} = \frac{cos<\mathcal{F}_v(v), \mathcal{F}_m(m)>}{cos<\mathcal{F}_t(t), \mathcal{F}_m(m)>},\label{eq:tv_ratio}
\end{align}
where the definitions of $\mathcal{F}_m, \mathcal{F}_v, \mathcal{F}_t$ are given in Eq.~(\ref{eq:video_encoder}). 
$R_{vt}$ is the ratio between the cosine similarity of vision-video and that of text-video and measures the extent of modality bias of the multi-modal encoder $\mathcal{F}_m$.

For the base model in Eq.~(\ref{eq:base}), we compute $R_{vt}$ for a randomly sampled set of videos and plot the density histogram graph in Fig.~\ref{subfig:rtv}. As observed, most $R_{vt}<0.3$, indicating the model suffer from the modality imbalance problem and the multimodal embeddings are heavily biased to the text contents. Consequently, when retrieving videos with the base model, visually irrelevant videos can be recalled falsely with even higher similarities than videos relevant to the queries textually and visually. 
\textit{The fundamental cause of modality imbalance is that text matching provides a \textbf{shortcut} for the bi-encoder: the query-text relation is easier to grasp, and most samples in training set can be solved by lexical relevance.}

\subsection{Modality-Shuffled Negatives}
\label{subsec:shuf}

To eliminate the shortcut, we generate novel \textbf{M}odality-\textbf{S}huffled (\textbf{MS} for short) negative samples, whose vision modality is irrelevant with the query, while the text is relevant to the query. As illustrated in fig.~\ref{subfig:sim_density_bi}, such MS negatives are serious adversarial attacks for the base model, which cannot distinguish MS negatives with the real positives. This inspires our loss design of MS negatives as below:
 \begin{align}   
 \mathcal{L}_{ms} = -\log\Big(\frac{\exp(s(q, m)/\tau)}{\exp(s(q, m)/\tau)+\sum_{\hat{m}\in\mathcal{M}_{ms}}\exp(s(q, \hat{m})/\tau)}\Big), \label{eq:ms-pos-diff}
\end{align}
where $\mathcal{M}_{ms}$ denotes the set of generated MS negatives.

$\mathcal{L}_{ms}$ straightly promotes the model disentangle the MS negatives from the real positives as shown in Fig.~\ref{subfig:sim_density}. \textit{By the way, it is not hard to find that when both $R_{vt}$ and $R_{\hat{v}t}$ are close to 0, $\mathcal{F}_{m}(t, v)$ will be close to its MS negative $\mathcal{F}_{m}(t, \hat{v})$.}
Thus, \modelname~ with $\mathcal{L}_{ms}$ also pushes $R_{vt}$ faraway from 0 as in Fig.~\ref{subfig:rtv}, which indicates that $\mathcal{L}_{ms}$ effectively alleviates the modality imbalance problem. Consequently, the information of both text and vision can be well-preserved in the final video embedding. 

\textbf{How to generate MS negatives efficiently?} We design to re-combine text embeddings and vision embeddings \textit{in the mini-batch} as in Fig.~\ref{fig:main}. For a mini-batch of size $n$, the vision, text embeddings of the $k$-th video are $\mathcal{F}_v(v_k)$, $\mathcal{F}_t(t_k)$. Then the MS negatives of the k-th video can be computed as $\mathcal{H}(\mathcal{F}_v(v_l), \mathcal{F}_t(t_k))$, where $l$ is a randomly selected integer from $1$ to $n$ except $k$. \textbf{Such design can generate one MS negative for each video with only one MSA operation, which is extremely efficient.} By repeating the above process $M$ times, we can generate $M$ MS negatives for each video. Empirically, more MS negatives can result in better performance, we set M as 32 to balance the effectiveness and efficiency.

\subsection{Dynamic Margin} 
\label{subsec:dynamic}
 To further address the modality bias, we apply a dynamic margin $\lambda$ on the positive pair $(q, m)$  of $\mathcal{L}_{qm}$ as below:
\begin{small}
\begin{align}
    \mathcal{L}_{qm} = -\log\Big(\frac{\exp((s(q, m)-\lambda)/\tau)}{\exp((s(q, m)-\lambda)/\tau)+\sum_{\hat{m}\in\mathcal{M}^-_q}\exp(s(q, \hat{m})/\tau)}\Big) \label{eq:qm_margin}.
\end{align}
\end{small}
$\lambda$ is computed from the visual relevance of $(q, m)$ through a scale and shift transformation:
\begin{align}
    \lambda = w\sigma(cos<\mathcal{F}_v(v), \mathcal{F}_q(q)>)+b, \label{eq:margin}
\end{align}
where $w=0.3, b=-0.1$, and $\sigma$ denotes the sigmoid function, \ie, $\sigma(x)=\frac{1}{1+e^{-x}}$. Then the margin $\lambda$ varies in $(-0.1, 0.2)$ and monotonically increases \wrt~ the visual relevance (\ie, $cos<\mathcal{F}_v(v), \mathcal{F}_q(q)>$). We also do the same modification for the video-to-query loss $\mathcal{L}_{mq}$ and the MS loss $\mathcal{L}_{ms}$. Then the main objective $\mathcal{L}_{bi}$ in Eq.~\ref{eq:bi} with dynamic margin is referred as $\widetilde{\mathcal{L}}_{bi}$ and $\mathcal{L}_{ms}$ with dynamic margin as $\widetilde{\mathcal{L}}_{ms}$. Note that the gradient of the margin $\lambda$ is detached during model training. 

To understand the effect of the dynamic margin easily, when $\lambda>0$, it can be considered as moving the positive video $m$ a bit faraway from the query before computing the loss and results in a larger loss value as in Fig.~\ref{fig:main}. Therefore, the dynamic margin encourages the model to produce even higher similarities for the vision related query-video pairs. 
 
Finally, the overall learning objective of our \modelname~ framework is as follows,
\begin{align}
    \mathcal{L} =  \widetilde{\mathcal{L}}_{bi} + \alpha\mathcal{L}_{v} +  \beta\mathcal{L}_{t} + \gamma\widetilde{\mathcal{L}}_{ms}, \label{eq:m}
\end{align}
where $\gamma$ is a weight hyper-parameter and set as $0.01$. In summary, \modelname~ solves the modality imbalance issue from two collaborative aspects: punishing on videos with unrelated vision contents with the MS component, and enhancing query-video pairs of related vision contents with the dynamic margin. 

\pdfoutput=1
\section{Experiments}
\label{sec:exp}

\subsection{Experimental Setup}
In this section, we describe our datasets, evaluation metrics.

\vpara{Training Dataset.} 
The training dataset contains about 5 million queries, 42 million videos and 170 million relevant query-video pairs mined from recent nine months' search logs.  

\vpara{Evaluation Datasets.} 
We do offline evaluations on two test datasets. \mandata: a manually annotated dataset of two million query-video pairs to evaluate the relevance of recalled videos; \autodata: a dataset of five million randomly sampled query-video pairs that automatically collected with Wilson CTR from search logs to simulate the online performance. The datasets used in the paper will be public only as embedding vectors to protect the privacy of users. 

For all models, we compute their Precision@K and MRR@K on \autodata and PNR on \mandata, which are defined as below:

\vpara{Precision@K.} Given a query $q$, $\mathcal{V}_q$ is the set of relevant videos, and the top $K$ documents returned by a model is denoted as $\mathcal{R}_q = \{r_1, \cdots, r_K\}$. The metric of Precision@$K$ is defined as
\begin{small}
\begin{align}
Precision@K = \frac{|\mathcal{R}_q \cap \mathcal{V}_q|}{K}.
\end{align}
\end{small}
\vpara{MRR@K.} Mean Reciprocal Rank at K (MRR@K) is defined as
\begin{small}
\begin{align}
MRR@K = \frac{1}{K}\sum_{i=1}^K\mathcal{I}_{r_{i} \in \mathcal{V}_q}\cdot\frac{1}{i},
\end{align}
\end{small}
where $\mathcal{I}_{\mathcal{A}}$ is an indicator function\footnote{Compared with Precision@K, MRR@K can reflect the order of top-$K$. Note that MRR@K is usually computed when there is only one positive sample and its value is always below 1, whereas we have more than one relevant videos for each query, then the value of MRR@K can be larger than 1.}. If $\mathcal{A}$ holds, it is 1, otherwise 0. 


\vpara{PNR.} 
For a given query $q$ and its associated videos $\mathcal{D}_q$, the positive-negative ratio (PNR) can be defined as
\begin{small}
\begin{align}
PNR = \frac{\sum_{d_i,d_j\in \mathcal{D}_q} \mathcal{I} (y_i > y_j)\cdot \mathcal{I} (s(q,d_i) > s(q,d_j))}{\sum_{d_{i'},d_{j'}\in \mathcal{D}_q} \mathcal{I} (y_{i'} > y_{j'})\cdot \mathcal{I} (s(q,d_{i'}) < s(q,d_{j'}))},
\end{align}
\end{small}
where $y_i$ represents the manual label of $d_i$, and $s(q, d_i)$ is the predicted score between $q$ and $d_i$ given by model. PNR measures the consistency between labels and predictions. 

\subsection{Offline Evaluations}
\label{subsec:offline}

\begin{table}[!t]
\centering
  \caption{
  \label{tab:comparsion}
    Offline experimental results of compared methods on \autodata and \mandata test sets.
  }
  \vspace{-0.1in}
{
\small
\begin{tabular}{lccc}
\toprule
 &\multicolumn{2}{c}{\textbf{Auto}} & \textbf{Manual} \\
Method & MRR@10 & Precision@10 (\%) & PNR 
\\ \midrule
\emph{Text} & 1.406 & 45.58 & 2.188\\ 
\emph{Vision} & 0.603 & 17.55 & 1.942\\ 
\emph{Base model} & 1.446 & 46.53  & 2.230\\ 
\midrule
\emph{+IAT\cite{lamb2019interpolated}} & 1.452  & 46.22  & 2.243 \\ 
\emph{+CDF\cite{kim2020modality}}  & 1.463  & 47.31  & 2.253 \\ 
 \midrule
\emph{+MS} & 1.600 &50.52  &\textbf{2.311} \\ 
\emph{+DM} & 1.491 & 47.75 & 2.267  \\ 
\emph{\modelname} & \textbf{1.614} & \textbf{51.10}  & 2.310 \\ 
\bottomrule
\end{tabular}
}
\end{table}

\vpara{Compared Methods}
We compare our method with the highly optimized baseline, and recent state-of-the-art modality-balanced techniques of IAT ~\cite{lamb2019interpolated} and CDF~\cite{kim2020modality}. \emph{IAT} trains a robust model under the adversarial attacks of PGD~\cite{madry2018towards} and \emph{CDF} aims to retrieve videos with one modality missed. 
\begin{itemize}[leftmargin=*]
    \item \emph{Base} is Eq.~\ref{eq:base} without MS negatives and dynamic margin.
    \item \emph{Text}~(\emph{Vision}) only uses the text (vision) modality of videos.
    \item \emph{+IAT} equips \emph{Base} with IAT~\cite{lamb2019interpolated}.
    \item \emph{+CDF} equips \emph{Base} with CDF~\cite{kim2020modality}.
    \item \emph{+MS} equips \emph{Base} with MS negatives $\mathcal{L}_{ms}$.
    \item \emph{+DM} equips \emph{Base} with dynamic margin $\widetilde{\mathcal{L}}_{bi}$.
    \item \emph{MBVR} is the full model with both \textit{DM} and \emph{MS}.
\end{itemize}

Table~\ref{tab:comparsion} illustrates experimental results of compared methods on both \autodata and \mandata test sets. The drastic performance difference between \emph{Vision} and \emph{Text} results from the dominated role of text modality in  video's modalities. \emph{+IAT} and \emph{+CDF} bring marginal improvements over \emph{Base}. 
Both \emph{+MS} and \emph{+DM} boost significantly over the strong baseline \emph{Base}. \emph{MS} is extremely effective, which brings nearly 4\% absolute boost over \emph{Base} ($46.53\%\longrightarrow50.52\%$). Furthermore, the full model \emph{MBVR}, with both \emph{MS} and \emph{DM}, achieves the best performance. The offline evaluations verify the effectiveness and compatibility of both components of ~\modelname~(\emph{MS} and \emph{DM}).

\subsection{Online Evaluations}
\label{subsec:online}
For the online test, the control baseline is current online search engine, which is a highly optimized system with multiple retrieval routes of text embedding based ANN retrieval, text matching with inverted indexing,~\etc, to provide thousands of candidates, and several pre-ranking and ranking models to rank these candidates. And the variant experiment adds our \modelname~ multimodal embedding based retrieval as an additional route. 

\vpara{Online A/B Test} We conducted online A/B experiments over 10\% of the entire traffic for one week. The watch time has increased by1.509\%; The long watch rate has increased by 2.485\%; The query changed rate\footnote{The decrease of query changed rate is positive, as it means the users find relevant videos without changing the query to trigger a new search request.} has decreased by 1.174\%.
 This is a statistically significant improvement and verifies the effectiveness of MBVR. Now \modelname~ has been deployed online and serves the main traffic.

\vpara{Manual Evaluation}
We conduct a manual side-by-side comparison on the top-4 videos between the baseline and the experiment. We randomly sample 200 queries whose top-4 videos are different, and then we let several human experts to tell whether the experiment's results are more relevant than the baseline ones. The Good \emph{vs.} Same \emph{vs.} Bad (GSB) metric is  G=45, S=126, B=29, where G (or B) denotes the number of queries whose results in experiments are more relevant (or irrelevant) than the baseline. 
This GSB result indicates that \modelname~ can recall more relevant videos to further meet the users' search requests.

\pdfoutput=1
\section{Conclusions and Discussions}
\label{sec:conclusion}
In this paper, we identify the challenging modality bias issue in multimodal embedding learning based on online search logs and propose our solution \modelname.
The main contributions of \modelname~ are the modality-shuffled (MS) negatives and the dynamic margin (DM), which force the model to pay more balanced attention to each modality. Our experiments verify that the proposed \modelname~ significantly outperforms a strong baseline and recent modality balanced techniques on offline evaluation and improves the highly optimized online video search system. 

As an early exploration of building multimodal retrieval system for short-video platforms, our \modelname~ adopts a succinct scheme (\ie, we keep engineering/architecture design choices simple). There are potential directions to further enhance the system,~\eg, using more frames, designing more sophisticated cross-modal fusion modules, and adopting smarter data cleaning techniques, which can be explored in the future.

\appendix
\section*{Acknowledgments}
We thank Xintong Han for early discussions of \modelname. We thank Xintong Han, Haozhi Zhang, Yu Gao, Shanlan Nie for paper review. We also thank Tong Zhao, Yue Lv for preparing training datasets for the experiments.


\bibliographystyle{ACM-Reference-Format}
\balance
\bibliography{references}

\end{document}